# Non-Singular Assembly-mode Changing Motions for *3-RPR* Parallel Manipulators

Mazen ZEIN, Philippe Wenger and Damien Chablat

Institut de Recherche en Communications et Cybernétique de Nantes UMR CNRS 6597

1, rue de la Noë, BP 92101, 44312 Nantes Cedex 03 France

E-mail address: Mazen.Zein@irccyn.ec-nantes.fr

**Abstract**

*When moving from one arbitrary location at another, a parallel manipulator may change its assembly-mode without crossing a singularity. Because the non-singular change of assembly-mode cannot be simply detected, the actual assembly-mode during motion is difficult to track. This paper proposes a global explanatory approach to help better understand non-singular assembly-mode changing motions for 3-RPR planar parallel manipulators. The approach consists in fixing one of the actuated joints and analyzing the configuration-space as a surface in a 3-dimensional space. Such a global description makes it possible to display all possible non-singular assembly-mode changing trajectories.*

## 1. Introduction

Most parallel manipulators have singularities that limit the motion of the moving platform. The most dangerous ones are the singularities associated with the direct kinematics, where two assembly-modes coalesce. Indeed, approaching such a singularity results in large actuator torques or forces, and in a loss of stiffness. Hence, these singularities are undesirable. There exists three main ways of coping with singularities, which have their own merits. A first approach consists in eliminating the singularities at the design stage by properly determining the kinematic architecture, the geometric parameters and the joint limits [1-2]. This approach is safe but difficult to apply in general and restricts the design possibilities. A second approach is the determination of the singularity-free regions in the workspace [3-7]. This solution does not involve a priori design restrictions but, because of the complexity of the singularity surfaces, it may be difficult to determine definitely safe regions. Finally, a third way consists in planning singularity-free trajectories in the manipulator workspace [8-11]. With this solution one is also faced with the complexity of the singularity equations but larger zones of the workspace may be exploited. This paper addresses a feature that has drawn the interest of quite few researchers but yet may concern many parallel manipulators, even planar ones [12, 13]: the fact that the manipulator can change its assembly-mode without passing through a singularity. Planar parallel manipulators with three extensible leg rods, referred to 3-*RPR*[1], have received a lot of attention because they have interesting potential applications in planar motion systems [12-18]. As shown in [13], moreover, the study of the 3-*RPR* planar manipulator may help better understand the kinematic behavior of its more complex spatial counterpart, the 6-dof octahedral manipulator. Planar parallel manipulators may have up to six assembly-modes [14]. It was first pointed out that to move from one assembly-mode to another, a 3-*RPR* planar parallel manipulator should cross a singularity [14]. But [12] showed, using numerical experiments, that this statement is not true in general. In fact, an analogous phenomenon exists in serial manipulators, which can move from one inverse kinematic solution to another without meeting a singularity [12-14]. The non-singular

---

[1] R and P stand for Revolute and Prismatic, respectively. The underlined letter refers to the actuated joint

change of posture in serial manipulators was shown to be associated with the existence of points in the workspace where three inverse kinematic solutions meet, called cusp points [19]. On the other hand, McAree and Daniel [13] pointed out that a 3-*RPR* planar parallel manipulator can execute a non-singular change of assembly-mode if a point with triple direct kinematic solutions exists in the joint space. The authors established a condition for three direct kinematic solutions to coincide and showed that a non-singular assembly-mode changing trajectory in the joint space should encircle a cusp point. Wenger and Chablat [20] investigated the question of whether a change of assembly-mode must occur or not when moving between two prescribed poses in the workspace. They defined the uniqueness domains in the workspace as the maximal regions associated with a unique assembly-mode and proposed a calculation scheme for 3-*RPR* planar parallel manipulators using octrees. They showed that up to three uniqueness domains exist in each singularity-free region. When the starting and goal poses are in the same singularity-free region but in two distinct uniqueness domains, a non-singular change of assembly-mode is necessary. However they did not investigate the kind of motion that arises when executing a non-singular change of assembly-mode. For the particular case of a planar 3-*RPR* parallel manipulator with similar base and platform, Kong and Gosselin [4] showed that there is no need to investigate non-singular assembly changing motion because each singularity-free region corresponds to one uniqueness domain. But 3-*RPR* manipulators with similar base and platform have a major flaw: the manipulator is singular at all positions when the moving platform assumes a zero orientation . On the other hand, planar 3-*RRR* manipulators and spatial octahedral manipulators with similar base and platform may change their assembly-mode without encountering a singularity [13, 21]. If, when moving from one arbitrary pose to another, the manipulator changes its assembly-mode without crossing a singularity, the actual assembly-mode during motion is difficult to track even if the initial assembly-mode is known, as there is no ways to detect the change of assembly-mode. Therefore, there is a need to understand the non-singular change of assembly-mode. The main goal of this paper is to investigate the non-singular change of assembly-mode in planar 3-*RPR* parallel manipulators, and to propose an explanatory approach to plan non-singular assembly-mode changing trajectories. The approach consists in fixing one of the actuated joints and analyzing the configuration-space as a surface in a 3-dimensional space. Such a global description makes it possible to display all possible non-singular assembly-mode changing trajectories.

## 2. Preliminaries

### 2.1 Manipulators Under Study

Figure1 shows a planar parallel manipulator with three extensible leg rods. The geometric parameters are the three sides of the moving platform $d_1$, $d_2$, $d_3$ and the position of the base revolute joint centers defined by $A_1$, $A_2$ and $A_3$. The reference frame is centered at $A_1$ and the $x$-axis passes through $A_2$. Thus, $A_1 = (0, 0)$, $A_2 = (A_{2x}, 0)$ and $A_3 = (A_{3x}, A_{3y})$. The joint space $Q$ is defined by the joint vectors $\mathbf{q}$ defined by the lengths of the three actuated extensible links: $\mathbf{q} = \begin{bmatrix} \rho_1 & \rho_2 & \rho_3 \end{bmatrix}^{\mathrm{T}}$. The task space is usually defined by the set of vectors $\mathbf{x} = \begin{bmatrix} x & y & \alpha \end{bmatrix}^{\mathrm{T}}$ where $(x, y)$ are the Cartesian coordinates of one point of the platform in the plane, chosen as $B_1$ in this paper, and $\alpha$ is the orientation of the platform in the plane with respect to the x-axis. In this paper, the task space will be more conveniently defined by $\mathbf{x} = \begin{bmatrix} \rho_1 & \theta_1 & \alpha \end{bmatrix}^{\mathrm{T}}$ where $(\rho_1, \theta_1)$ are the cylindrical coordinates of $B_1$. With these parameters, indeed, it is possible to consider 2-dimensional slices of the joint space and of the workspace by fixing the joint parameter $\rho_1$.

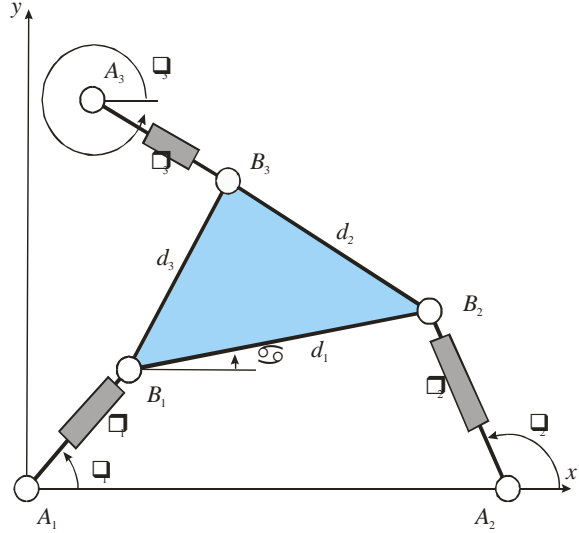

**Figure 1: The 3-*RPR* parallel manipulator under study**

To illustrate our work, we refer to the same *3-RPR* manipulator as the one used in [12, 13,15], which has the following geometric parameters: $A_1$= (0, 0), $A_2$= (15.91,0), $A_3$ = (0, 10), $d_1$= 17.04, $d_2$= 16.54 and $d_3$ = 20.84 in an arbitrary length unit.

## 2.2    Kinematic relations

The relation between the joint space **Q** and the output space can be expressed as a system of non-linear algebraic equations, which can be written as:

$$F(\mathbf{x}, \mathbf{q}) = 0 \qquad (1)$$

Differentiating equation (1) with respect to time leads to the velocity model:

$$\mathbf{At} + \mathbf{B\dot{q}} = \mathbf{0}$$

where $\mathbf{t} = \begin{bmatrix} \omega, \dot{\mathbf{c}} \end{bmatrix}^T$, $\omega$ is the scalar angular velocity and $\dot{\mathbf{c}}$ is the two-dimensional velocity vector of the operational point $B_1$ of the platform. **A** and **B** are 3×3 Jacobian matrices which are configuration dependent, and $\dot{\mathbf{q}} = \begin{bmatrix} \dot{\rho}_1 & \dot{\rho}_2 & \dot{\rho}_3 \end{bmatrix}^T$ is the joint velocity vector.

## 2.3    Singular configurations

The singularities of the 3-*RPR* planar parallel manipulators have been extensively studied [13, 16, 22, 24, 25]. On a singular configuration of the manipulator, matrix **A** or matrix **B** or both of them are singular. In this study, only the singularities of **A** are of interest.

To derive the singularity equations, it is usual to expand the determinant of **A**. We use rather a geometric approach that does not involve complicated algebraic calculus. The *3-RPR* parallel manipulator is on a singular configuration whenever the axes of its three legs are concurrent or parallel [24] (Fig. 2).

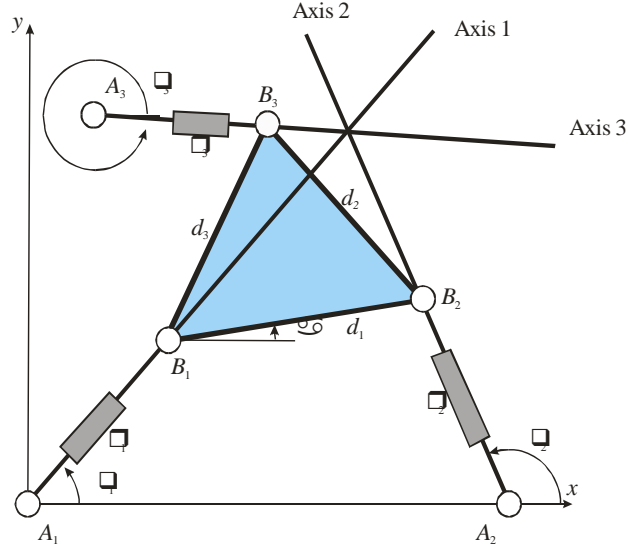

**Figure 2: A 3-*RPR* parallel manipulator in a singular configuration**

The equations of the three leg axes can be written as:

$$\begin{cases} (\text{Axis 1}) : y\cos(\theta_1) = x\sin(\theta_1) \\ (\text{Axis 2}) : y\cos(\theta_2) = (x - A_{2x})\tan(\theta_2) \\ (\text{Axis 3}) : y\cos(\theta_3) = (x - A_{3x})\sin(\theta_3) + A_{3y}\cos(\theta_3) \end{cases} \tag{2}$$

The condition of these three axes to intersect (possibly at infinity) is:

$$A_{2x}\sin(\theta_2)\sin(\theta_3 - \theta_1) + \big(A_{3x}\sin(\theta_3) - A_{3y}\cos(\theta_3)\big)\sin(\theta_1 - \theta_2) = 0 \tag{3}$$

which is the singularity equation of the manipulator. This expression along with the constraint equations of the manipulator (i.e. writing the fixed distances between the three vertices of the mobile platform $B_1$, $B_2$, $B_3$) allow us to plot the singular curves in 2-dimensional slices of the joint space ($\rho_2$, $\rho_3$) and of the workspace ($\alpha$, $\theta_1$) for a fixed value of $\rho_1$ [13, 26].

### 2.3.1   *Workspace singularities*

Figure 3a shows the singular curves in the workspace slice ($\alpha$, $\theta_1$) defined by $\rho_1$=17 for the manipulator introduced in Section **2.1**. Note that because the space ($\alpha$, $\theta_1$) is a torus (the revolute joints are assumed unlimited) the opposite sides of the square representation in Fig. 3a are actually coincident. Thus, the singularity curves divide the workspace into two connected components called *aspects* [7]. The notion of aspects was first introduced for serial manipulators by [27] to cope with the existence of multiple inverse kinematic solutions. The aspects were defined as the maximal singularity-free domains in the joint space. The aspects were extended in [7] to parallel manipulators with only one inverse kinematic solution such as *3-RPR* manipulators. For such manipulators, the aspects are the maximal singularity-free connected regions in the workspace. An equivalent definition was used in [28] for a special case of parallel manipulators.

For the *3-RPR* parallel manipulator at hand, the first (resp. second)  aspect is defined by det(**A**)>0 (resp. det(**A**)<0), where **A** is the Jacobian matrix introduced in Section **2.2**.

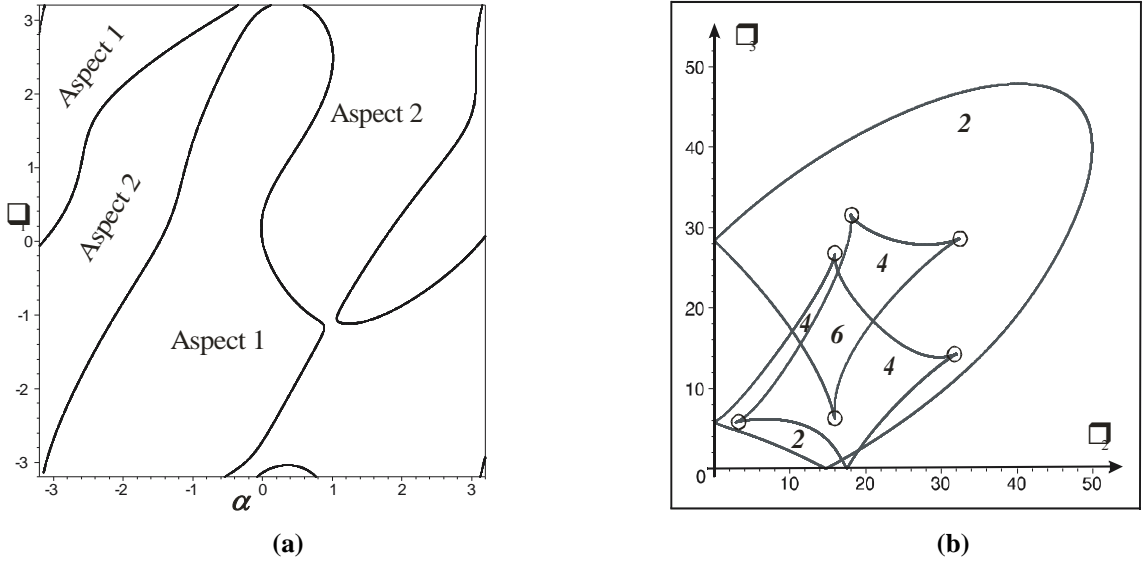

**(a)**                                                **(b)**

**Figure 3:** Singular curves in ($\alpha$, $\theta_1$) for $\rho_1$=17 (Fig. 3a). Singular curves in ($\rho_2$, $\rho_3$) for $\rho_1$=17 (Fig. 3b).

### 2.3.2   *Joint space singularities*

Figure 3b shows the singular curves in the joint space slice ($\rho_2$, $\rho_3$) defined by $\rho_1$=17. These curves split the joint space slice into several regions with 2, 4 or 6 direct kinematic solutions [26, 29]. The slice of the joint space is the image by the inverse kinematics of each of aspect 1 and also of aspect 2. That is, plotting either aspect 1 or aspect 2 onto ($\rho_2$, $\rho_3$) will define the pattern depicted in Fig. 3b. In each region, the number of direct solutions is equally distributed in the two aspects. In the central 6-solution region, for example, there are 3 solutions in aspect 1 and 3 solutions in aspect 2.

The six points pinpointed with circles are cusp points, where three direct kinematic solutions coincide. As shown in [26], there may be 0, 2, 4, 6 or 8 cusp points depending on the slice chosen.

## 3.   Examination of a loop trajectory encircling a cusp point in the joint space

We want to understand how a non-singular change of assembly-mode arises. We consider that no external influence enabling the determination of the assembly mode is possible. We only know that to execute a non-singular assembly-mode changing motion, a cusp point must be encircled in the joint space of the manipulator [13]. We show in this section that this information is insufficient to determine the actual motion of the manipulator when a cusp point is encircled in its joint space. Let us define a triangular loop trajectory $T$ in a slice of the joint space for $\rho_1 = 17$, which encircles a cusp point (Fig. 4a). The starting (and final) joint vector $\mathbf{q}=\begin{bmatrix}\rho_1 & \rho_2 & \rho_3\end{bmatrix}^T$ is chosen as $\mathbf{q}=\begin{bmatrix}17 & 19 & 17\end{bmatrix}^T$, where the direct kinematics at $\mathbf{q}$ admits 6 real solutions $P_i$, i=1, 2, …6. Keeping in mind that the opposite sides of the square representation in Fig. 4b are actually coincident, $P_1$, $P_2$ and $P_3$ are in aspect 1 and the remaining three solutions are in aspect 2. The loop trajectory $T$ crosses the singular curves at four distinct joint configurations, referred to as $\mathbf{q}_a$, $\mathbf{q}_b$, $\mathbf{q}_c$ and $\mathbf{q}_d$. Along $T$, the direct kinematics is solved and the solutions are plotted in ($\alpha$, $\theta_1$) (Fig. 4b). According to whether $T$ is executed clockwise or counter-clockwise and according to the initial assembly-mode, a total of 12 motions will result in the workspace. These 12 motions can be classified into the following three types:

- Motions that make the manipulator stop at one of the singular points $S_a$, $S_b$, $S_c$ or $S_d$, which are associated with $\mathbf{q}_a$, $\mathbf{q}_b$, $\mathbf{q}_c$ and $\mathbf{q}_d$, respectively. There are 8 such motions, 4 for each direction of execution of $T$. When $T$ is executed clockwise the 4 motions are from $P_2$ to $S_a$, from $P_1$ to $S_d$, from $P_5$ to $S_a$ and from $P_6$ to $S_d$. When $T$ is executed counter-clockwise the 4 motions are from $P_3$ to $S_b$, from $P_1$ to $S_c$, from $P_5$ to $S_b$ and from $P_6$ to $S_c$. In each case, the direct kinematic solution associated with the starting assembly-mode is lost at the singular point and this is the reason why the motion stops and $T$ cannot be fully executed. Thus, no assembly-mode changing is feasible with these motions.

- Two loop motions in the workspace starting and ending at $P_4$ (Fig. 4b). These two loops differ in their direction of execution (clockwise or counter-clockwise), which depends on the direction of execution of $T$. These motions do not enable the manipulator to change its assembly-mode because the moving platform goes back to its starting pose in the workspace. Unlike the 8 motions described above, for these two loop motions, $T$ is fully executed. These loop motions feature three segments that are associated with the three linear segments of the triangular trajectory $T$.

- Two non-singular motions that differ only in their direction of motion (from $P_2$ to $P_3$ or from $P_3$ to $P_2$, depending on the starting assembly-mode). The path associated with these motions is drawn in dark grey in Fig. 4b. Again, the path is composed of three segments, each associated with a segment of the triangular trajectory $T$. But in this case, the arrival assembly-mode is different from the starting one. Thus, these two motions are non-singular assembly-mode changing motions.

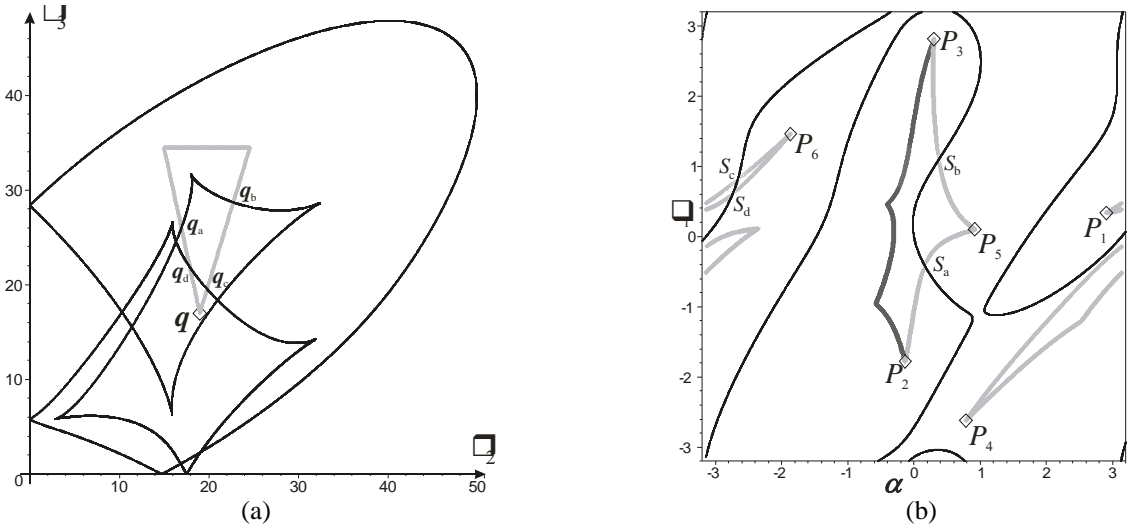

(a)                                    (b)

**Figure 4: A loop trajectory encircling a cusp point in the joint space section $\rho_1$=17 (Fig. 4a) and the associated motions in the workspace (Fig. 4b).**

Finally, this analysis raises the following questions or comments:

- Only two assembly-modes, namely $P_2$ and $P_3$, were found to be linkable by a non-singular assembly-mode changing motion. Yet, other non-singular motions should be found in aspect 1 (e.g. between $P_2$ and $P_1$ or between $P_3$ and $P_1$) and even in aspect 2.

- What would have been the resulting motions in the workspace if $T$ had encircled another cusp point?

- If one wants to connect two assembly-modes in the workspace associated with the same joint vector **q**, without crossing a singularity, which cusp point should be encircled?

Clearly, these questions cannot be answered with the sole information provided by the singularity locus in the joint space and in the workspace.

## 4.    A model for the configuration-space structure

### 4.1    The configuration-space as a surface in a 3-dimensional space

In the singularity loci shown in Fig. 4, we lose information as these loci result, in fact, from the projection into either $(\rho_1, \rho_2, \rho_3)$ or $(\rho_1, \alpha, \theta_1)$ of the configuration-space $CS$ of the manipulator, which is a 3-dimensional space embedded in the product space of $(\rho_1, \rho_2, \rho_3)$ and $(\rho_1, \alpha, \theta_1)$. When the first leg rod length is fixed, we consider a 2-dimensional slice of the configuration-space, which is thus a surface in a 4-dimensional space (the product space of $(\rho_2, \rho_3)$ and $(\alpha, \theta_1)$). Figure 4 represents its projection onto the planes $(\rho_2, \rho_3)$ and $(\alpha, \theta_1)$. But we cannot depict a surface in a 4-dimensional space.

Because non-singular assembly-mode motions occur only inside an aspect, a configuration-space should be built for each aspect: it is necessary to build $CS_1$ and $CS_2$, the configuration-space in aspects 1 and 2, respectively. In order to show $CS_1$ and $CS_2$, they should be displayed in $(\rho_2, \rho_3, \alpha)$ or in $(\rho_2, \rho_3, \theta_1)$ rather than in the product space of $(\rho_2, \rho_3)$ and $(\alpha, \theta_1)$. But by doing so one should verify that the third parameter is sufficient to describe fully the configuration of the manipulator in an aspect when arbitrary values of the three actuated joints are given. It has been shown recently that for some 3-_RPR_ manipulators, called degenerate manipulators, two distinct assembly-modes are always associated with the same $\alpha$ and these two assembly-modes may lie in the same aspect [31]. On the other hand, no manipulators exist that have always two distinct assembly-modes per aspect for the same value of $\theta_1$ [32]. Thus, it is possible to build $CS_1$ and $CS_2$ in $(\rho_2, \rho_3, \theta_1)$ but not in $(\rho_2, \rho_3, \alpha)$.

### 4.2    Construction of the configuration-space in each aspect

Let us fix $\rho_1$ to the value corresponding to the current configuration of the manipulator. To build $CS_1$ and $CS_2$, the slice $(\rho_2, \rho_3)$ of the joint space is scanned and the parameter $\theta_1$ is determined by solving the characteristic polynomial in $\theta_1$ [31]. The sign of the determinant of the Jacobian matrix is determined. If this sign is positive, the resulting point $(\rho_2, \rho_3, \theta_1)$ is plotted in $CS_1$ otherwise it is plotted in $CS_2$. Figure 5 shows $CS_1$ and $CS_2$ when $\rho_1=17$. We have used a CAD interface to build a 3D mesh surface-plot and to enable rotating the viewing perspective. The singular curves, which define the boundaries of $CS_1$ and $CS_2$, have been displayed in bold lines. Their projections onto $(\rho_2, \rho_3)$ are shown in the figure. Assume the manipulator is in the configuration **q** shown in Fig. 4. The points associated with the six assembly-modes $P_1$, $P_2$,…, $P_6$ are shown (these points have been labeled as $P_1$, $P_2$,…, $P_6$ in Fig. 5 like in Fig. 4b for more simplicity). The first three are in aspect 1 while the last three are in aspect 2. This representation shows all feasible singularity-free motions between $P_1$, $P_2$ and $P_3$ as well as those between $P_4$, $P_5$ and $P_6$.

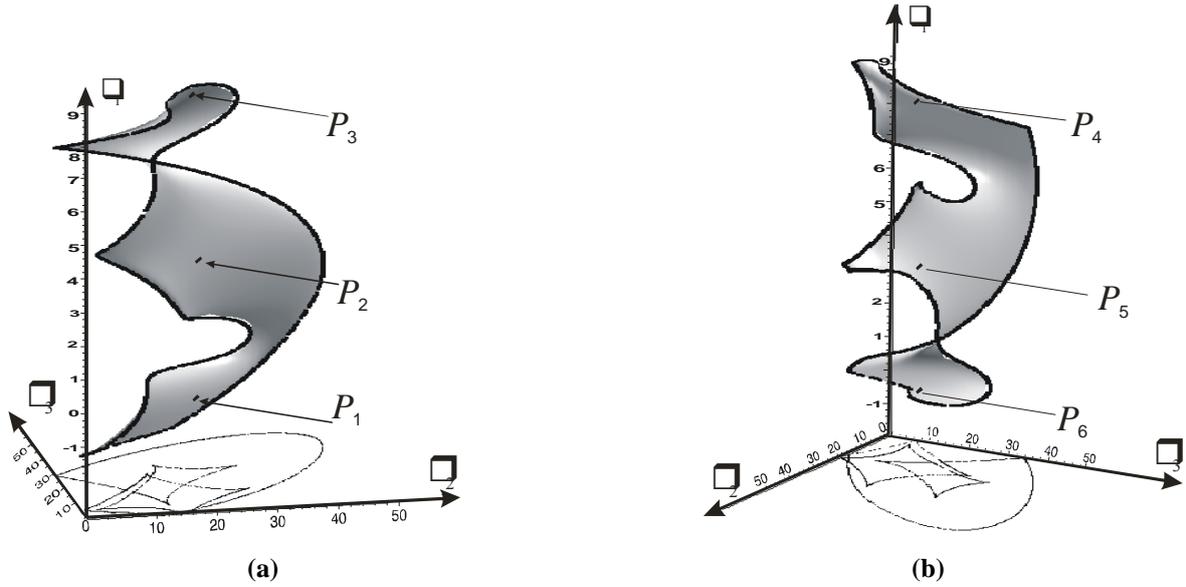

**(a)**                                                        **(b)**

**Figure 5: The configuration-space $CS_1$ (Fig. 5a) and $CS_2$ (Fig. 5b) associated with aspects 1 and 2, respectively, along with the six assembly-modes calculated at $\mathbf{q} = \begin{bmatrix} 17 & 19 & 17 \end{bmatrix}^T$ ( $P_1$, $P_2$ and $P_3$ in aspect 1 and $P_4$, $P_5$ and $P_6$ in aspect 2). The singularities (shown in bold lines) define the boundaries of $CS_1$ and $CS_2$.**

### 4.3   Examples of non-singular assembly-mode changing trajectories

The following three figures display non-singular assembly-mode changing trajectories in aspect 1. The paths are constructed in ($\rho_2$, $\rho_3$, $\theta_1$). We have used a CAD-interface to define the paths from a set of intermediate points and linear segments. In the three examples shown in Figs 6, 7 and 8, we have defined the path with three intermediate points. The projection of the paths onto ($\rho_2$, $\rho_3$) are displayed in each figure to show which cusp is encircled in the joint space. The viewpoints have been chosen to show as clearly as possible the paths in $CS_1$.

Figure 6 shows a non singular assembly-mode changing trajectory connecting the two assembly-modes $P_1$ and $P_2$. We notice that the encircled cusp point is different from the one encircled by the trajectory $T$ shown in Fig. 4a.

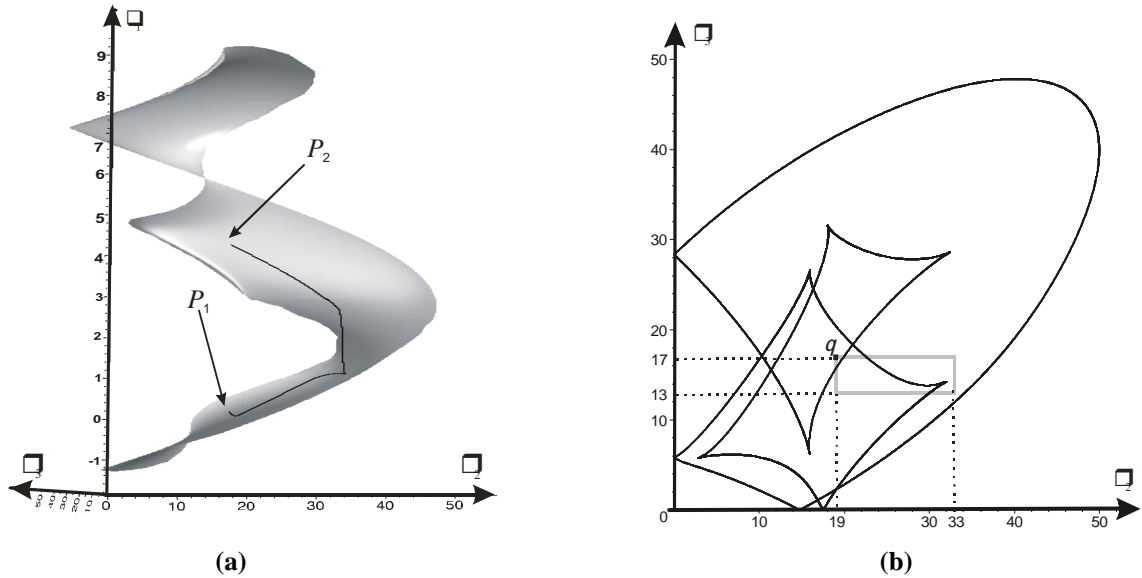

**(a)**                                                        **(b)**

**Figure 6: Non-singular assembly-mode changing trajectory connecting the two assembly-modes $P_1$ and $P_2$ in the first aspect in ($\rho_2$, $\rho_3$, $\theta_1$) (Fig. 6a), in ($\rho_2$, $\rho_3$) (Fig. 6b).**

Figure 7 shows a non-singular assembly-mode changing trajectory connecting the two assembly-modes $P_2$ and $P_3$. The trajectory encircles two cusp points but, in fact, only one needs to be encircled (namely, the one encircled by the trajectory *T* shown in Fig. 4, left).

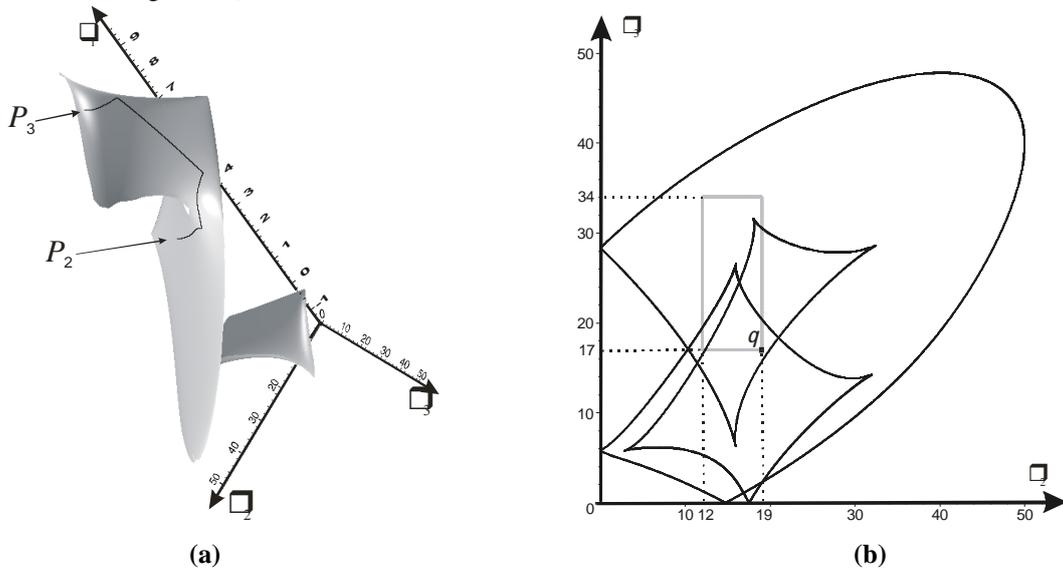

**(a)**                                                      **(b)**

**Figure 7: Non-singular trajectory connecting the two assembly-modes $P_2$ and $P_3$ in the first aspect in ($\rho_2$, $\rho_3$, $\theta_1$) (Fig. 6a), in ($\rho_2$, $\rho_3$) (Fig. 7b).**

Finally, Fig. 8 shows a non-singular assembly-mode changing trajectory connecting the two assembly-modes $P_1$ and $P_3$.

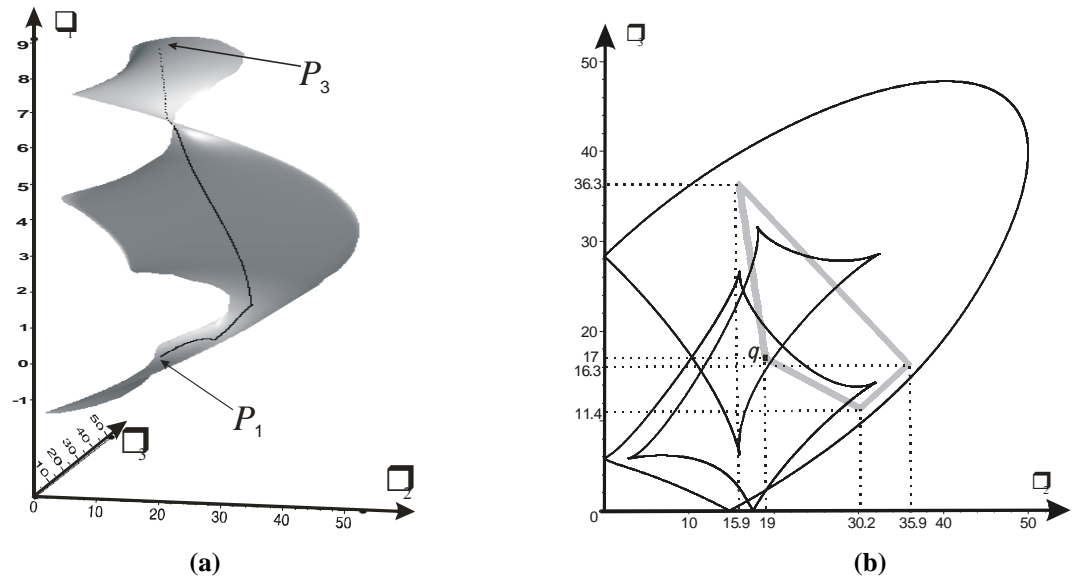

**(a)**                                                      **(b)**

**Figure 8: Non-singular trajectory connecting the two assembly-modes $P_1$ and $P_3$ in the first aspect in ($\rho_2$, $\rho_3$, $\theta_1$) (Fig. 8a), in ($\rho_2$, $\rho_3$) (Fig. 8b).**

We notice that to connect the assembly-modes $P_1$ and $P_3$, the trajectory in the joint space encircles two cusp points as shown in Fig. 8b. In fact, we have verified that any path from $P_1$ and $P_3$ must encircle these two cusp points. This fact can be explained by the "layered" structure of the configuration-space surfaces (see comments below). This is a new result, since it was thought that only one cusp point should be encircled when executing a non-singular assembly-mode changing maneuver [13]. Clearly, this result could not have been discovered with the sole joint space and workspace representations shown in Fig. 4.

## 4.4 Comments

Using the configuration-space surface model, the problem of which cusp point should be encircled to plan a non-singular assembly-mode changing motion needs not be solved. However, it is possible to know the response to this question by projecting onto $(\rho_2, \rho_3)$ the path built on $CS_1$ or $CS_2$.

The two configuration-space surfaces $CS_1$ and $CS_2$ may be regarded as being composed of three adjacent "layers" $L_1$, $L_2$ and $L_3$ with respect to the $\theta_1$-coordinate. These three layers are associated with the three assembly-modes $P_1$, $P_2$ and $P_3$. The first two layers and the last two layers are adjacent and it is possible to move between them with a smooth path. The presence of a cusp point in $(\rho_2, \rho_3)$ accounts for the existence of a continuous link between two layers of $CS_1$ or of $CS_2$. This is why moving from one layer to another (that is, moving from one assembly-mode to another without meeting a singularity) is equivalent to encircling a cusp point. This is because to move from $L_1$ to $L_3$, one has to go on the intermediate layer $L_2$ that two cusp points must be encircled in $(\rho_2, \rho_3)$ when connecting the two assembly-modes $P_1$ and $P_3$.

## 5. Conclusions

A global explanatory approach was proposed in this paper to help better understand non-singular assembly-mode changing motions for 3-$RPR$ planar parallel manipulators. It has been shown that the joint space and the workspace are not sufficient to describe the non-singular motions between assembly-modes. The proposed approach consists in fixing one of the actuated joint. Then the configuration-space is reduced to two surfaces in a 3-dimensional space, one for each aspect. Such a global description provides sufficient information as to the configuration space topology and makes it possible to display all possible non-singular assembly-mode changing trajectories that operate with one locked actuator.

This approach will be extended to the analysis of non-singular assembly-mode changing trajectories in 6-DOF octahedral parallel manipulators. These manipulators are particular Gough-Stewart platforms that feature a triangular base and a triangular moving platform, connected by six double-spherical-joint-ended rods. This manipulator has eight triangular faces (the base-triangle, the platform-triangle and 6 leg-triangles). The idea is to fix all but two joint coordinates and to build the configuration-space surfaces in $(\rho_5, \rho_6, \theta_1)$, where $\rho_5$ and $\rho_6$ are the two free joint coordinates and $\theta_1$ is the dihedral angle between the base and one of the leg-triangles.

We think that the work provided in this paper constitutes a first step to the difficult problem of how to identify the different assembly-modes of a parallel manipulator and how to track the assembly-mode during motion.